\documentclass{article}

\usepackage[preprint]{neurips_2024}

\usepackage[numbers,sort&compress]{natbib}
\usepackage{wrapfig}
\usepackage[utf8]{inputenc} % allow utf-8 input
\usepackage[T1]{fontenc}    % use 8-bit T1 fonts
\usepackage{hyperref}       % hyperlinks
\usepackage{url}            % simple URL typesetting
\usepackage{booktabs}       % professional-quality tables
\usepackage{amsfonts}       % blackboard math symbols
\usepackage{nicefrac}       % compact symbols for 1/2, etc.
\usepackage{microtype}      % microtypography
\usepackage{xcolor}         % colors
\usepackage{pifont}
\usepackage[pdftex]{graphicx}
\usepackage{multirow}
\usepackage{enumitem}

\title{Diffusion4D: Fast Spatial-temporal Consistent \\ 4D Generation via Video Diffusion Models}

\author{
  \textbf{Hanwen Liang\textsuperscript{1}\thanks{Equal contribution}\,, Yuyang Yin\textsuperscript{2$\ast$}, Dejia Xu\textsuperscript{3}, Hanxue Liang\textsuperscript{4}}, \\
  \textbf{Zhangyang Wang\textsuperscript{3}, Konstantinos N. Plataniotis\textsuperscript{1}, Yao Zhao\textsuperscript{2}, Yunchao Wei\textsuperscript{2}\textdagger}\\
  {\textsuperscript{1}University of Toronto}, {\textsuperscript{2}Beijing Jiaotong University},\\
  {\textsuperscript{3}University of Texas at Austin}, {\textsuperscript{4}University of Cambridge} 
  \\
  \\
  \small{\url{https://vita-group.github.io/Diffusion4D}}
}

\begin{document}

\maketitle
%\footnotetext[1]{Equal contribution.}
\renewcommand{\thefootnote}{\fnsymbol{footnote}}
\footnotetext[2]{Corresponding author.}
\renewcommand{\thefootnote}{\arabic{footnote}}

\begin{abstract}
  The availability of large-scale multimodal datasets and advancements in diffusion models have significantly accelerated progress in 4D content generation. 
  Most prior approaches rely on multiple image or video diffusion models, utilizing score distillation sampling for optimization or generating pseudo novel views for direct supervision.
  However, these methods are hindered by slow optimization speeds and multi-view inconsistency issues.
  Spatial and temporal consistency in 4D geometry has been extensively explored respectively in 3D-aware diffusion models and traditional monocular video diffusion models.
  Building on this foundation, we propose a strategy to migrate the temporal consistency in video diffusion models to the spatial-temporal consistency required for 4D generation.
  Specifically, we present a novel framework, \textbf{Diffusion4D}, for efficient and scalable 4D content generation. 
  Leveraging a meticulously curated dynamic 3D dataset, we develop a 4D-aware video diffusion model capable of synthesizing orbital views of dynamic 3D assets.
  To control the dynamic strength of these assets, we introduce a 3D-to-4D motion magnitude metric as guidance.
  Additionally, we propose a novel motion magnitude reconstruction loss and 3D-aware classifier-free guidance to refine the learning and generation of motion dynamics.
  After obtaining orbital views of the 4D asset, we perform explicit 4D construction with Gaussian splatting in a coarse-to-fine manner. 
  The synthesized multi-view consistent 4D image set enables us to swiftly generate high-fidelity and diverse 4D assets within just several minutes.
  Extensive experiments demonstrate that our method surpasses prior state-of-the-art techniques in terms of generation efficiency and 4D geometry consistency across various prompt modalities. 
  % Project page: \url{https://vita-group.github.io/Diffusion4D}
  % This innovative approach not only addresses the limitations of previous methods but also sets a new benchmark for 4D content generation.
\end{abstract}

\section{Introduction}
\label{sec:intro}
The availability of internet-scale image-text-video datasets, along with the progress in diffusion model techniques~\cite{song2020denoising,ruiz2023dreambooth,rombach2022high}, has propelled significant advancements in generating diverse and high-quality visual content, including images, videos, and 3D assets~\cite{nichol2021glide, ho2022imagen, liu2023zero, shi2023mvdream}. 
These advancements have further fostered much progress in the realm of 4D content generation~\cite{xu2024comp4d,jiang2023consistent4d,yin20234dgen,bahmani20234d, gao2024gaussianflow, ren2023dreamgaussian4d, singer2023text,chu2024dreamscene4d}, which has gained widespread attention across both research and industrial domains. The ability to generate high-quality 4D content is key to various applications, ranging from the artistic realms of animation and film production to the dynamic worlds of augmented reality. 
% \hx{, which recently has gained widespread attention across both research and industrial domains. The ability to generate high-quality 4D content is key to various applications, ranging from the artistic realms of animation and film production to the dynamic worlds of gaming and the burgeoning Metaverse.}
% 4D generation has gained widespread attention across both research and industrial domains, as the ability to generate high-quality 4D content is key to various applications, ranging from the artistic realms of animation and film production to the dynamic worlds of gaming and the burgeoning Metaverse.
% Along with the breakthroughs in dynamic 3D (4D) representation, these advancements further foster much improvement in the realm of 4D generation. This endevour receives much attention across research and industrial fields, as high-quality 4D content generation is of great value to a wide range of applications in animation, film, game and MateVerse.

\begin{figure}
    \begin{center}
        \includegraphics[width=1.0\linewidth]
        {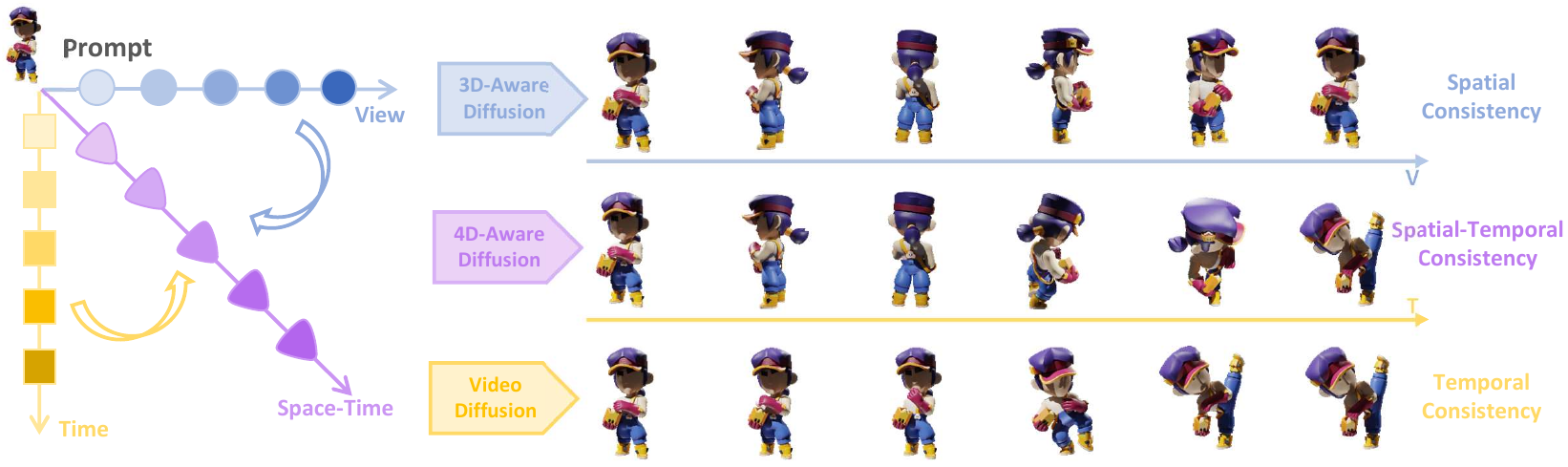}
        \caption{Decomposition of spatial-temporal consistency in 4D generation. The proposed Diffusion4D embeds geometrical consistency and temporal coherence into a single network.}
        \label{fig:intro}
    \end{center}
    \vspace{-0.5cm}
\end{figure}

%%% To point out existing problems (slow optimization, inconsistency) %%%
However, generating high-quality 4D content \textbf{efficiently} and \textbf{consistently} remains a significant challenge.
% Earlier 4D synthesis works~\cite{bahmani20234d,zhao2023animate124,singer2023text} primarily focused on generating 4D content conditioned on text or images. 
\ding{202} Due to the scarcity of large-scale multi-view consistent 4D datasets, earlier 4D synthesis works~\cite{bahmani20234d,zhao2023animate124,singer2023text} borrow appearance and motion priors from pretrained image- or video-diffusion models, and leverage score distillation sampling (SDS)~\cite{poole2022dreamfusion} for optimization. 
This strategy is time-consuming and computationally inefficient due to the heavy supervision back-propagation, limiting its widespread applicability. 
~\cite{yin20234dgen,zeng2024stag4d} explored video-conditioned 4D generation and utilized 3D-aware image diffusion models to acquire pseudo multi-view data for photometric supervision. Yet the issue of slow optimization speed persists.
\ding{203} Another primary challenge in 4D synthesis is to guarantee 4D geometry consistency, which can be decomposed into two components: spatial consistency and temporal consistency. As illustrated in Fig.~\ref{fig:intro}, spatial consistency implies that the object maintains consistent 3D geometry at each distinct timestamp, while temporal consistency indicates that the object's appearance and movement exhibit coherence and smoothness across timestamps. 
These two components have been thoroughly explored separately in static multi-view 3D synthesis~\cite{liu2023zero,tang2024mvdiffusion++,shi2023mvdream} and monocular video generation~\cite{blattmann2023stable, singer2022make, blattmann2023align}. 
Multi-view 3D diffusion models possess robust knowledge of geometrical consistency, whereas video diffusion models encapsulate temporal appearance coherence and smooth motion priors. 
Recent approaches~\cite{pan2024fast,yang2024diffusion} tried combining these two generative models to enhance efficiency and consistency in 4D generation. 
However, their underlying assumption of spatial-temporal conditional independence, the utilization of multi-model supervision, or inference at distinct timestamps inherently leads to 4D geometry inconsistencies.
This raises the question: \textbf{can we integrate spatial and temporal consistency into a single network, and obtain multi-timestamp cross-view supervision in one shot?}
% or "Our motivation is to address these limitations by combining space-time consistency into a single network."
Inspired by recent works in static 3D generation~\cite{voleti2024sv3d,zuo2024videomv} that repurposed temporal consistency in video generation to spatial consistency in 3D generation, we design a strategy that achieves spatial-temporal 4D consistency with a singular 4D-aware video diffusion model.

To this end, we propose a novel framework, \textbf{Diffusion4D}, to achieve efficient and consistent generation of 4D content.
First, to overcome the data scarcity issue, we meticulously curate a large-scale, high-quality 4D dataset from a vast corpus of 3D datasets~\cite{deitke2023objaverse,deitke2024objaverse}. 
Utilizing the curated dataset, we develop a \textit{4D-aware video diffusion model} capable of generating orbital videos of dynamic 3D objects, imitating the photographing process of 4D assets.
To the best of our knowledge, this is the first endeavor to adapt a video diffusion model and train on a 4D dataset for explicit novel view synthesis of 4D assets.
To explicitly control the dynamic strength of the assets, we propose a novel 3D-to-4D motion magnitude metric and embed it as conditions into the diffusion network.
We also incorporate the motion magnitude reconstruction loss to encourage the model to directly learn the object's 3D-to-4D dynamics.
During the sampling stage, we propose a novel 3D-aware classifier-free guidance to further enhance the dynamics of 3D assets.
Leveraging the flexibility of the video diffusion model architecture, our framework can seamlessly accommodate various prompts, including text, single image, and static 3D content.
It demonstrates the capability to generate highly spatial-temporal consistent orbital views around 4D assets, thus providing comprehensive information for the construction of 4D assets.
Building upon this foundation, we perform explicit \textit{4D construction} by optimizing Gaussian splatting representations~\cite{wu20234d} in a coarse-to-fine manner with photometric losses. 
The synthesized multi-view consistent 4D image set enables us to swiftly generate high-fidelity and diverse 4D assets within just several minutes.

In summary, our contributions can be summarized into three folds:
\begin{itemize}[itemsep=0em, topsep=0em, leftmargin=1em]
    \item We present a novel 4D content generation framework that, for the first time, adapts video diffusion models for explicit synthesis of spatial-temporal consistent novel views of 4D assets. The 4D-aware video diffusion model can seamlessly integrate with the off-the-shelf modern 4D construction pipelines to efficiently create 4D content. 
    %  \item We present \textbf{Diffusion4D}, a novel framework that achieves efficient and scalable 4D content generation.
    % Diffusion4D, for the first time, adapts video diffusion models for explicit synthesis of spatial-temporal consistent novel views of 4D assets. The 4D-aware video diffusion model trained on curated high-quality 4D dataset can seamlessly integrate with the off-the-shelf modern 4D construction pipelines to efficiently create 4D content. 
    \item We introduce the 3D-to-4D motion magnitude metric to enable explicit control over the dynamic strength and propose the motion magnitude reconstruction loss and 3D-aware classifer-free guidance to refine the dynamics learning and generation.
    \item Extensive experiments demonstrate that the proposed framework outperforms previous approaches in terms of generation efficiency and 4D geometry consistency, establishing a new benchmark for 4D generation under different types of prompts, including text, single image, and static 3D content.
\end{itemize}

% \vspace{-0.3cm}
\section{Background}
4D generation aims at synthesizing dynamic 3D content from various inputs, such as text, single images, monocular videos, or static 3D assets.
Extending 3D generation into the space-time domain, this task requires not only consistent geometry prediction but also the generation of temporally consistent dynamics.
There are many works dedicated to this task. 
MAV3D\cite{singer2023text} deals with text-conditioned 4D generation by utilizing SDS derived from video diffusion models to optimize a dynamic NeRF representation.
4DFY~\cite{bahmani20234d} tackles this problem by combining supervision signals from image, video, and 3D-aware diffusion models. 
Animate124~\cite{zhao2023animate124} leverages both text and image prompts to improve the appearance of the dynamic 3D assets. 
DreamGaussian4D \cite{ren2023dreamgaussian4d} adopts the deformable 3D Gaussians for 4D representations and tackles the image-conditioned generation with a driving video. 
Similarly, 4DGen~\cite{yin20234dgen} proposes utilizing spatial-temporal pseudo labels and obtains anchor frames with a multi-view diffusion model. 
Consistent4D \cite{jiang2023consistent4d} leverages the object-level 3D-aware image diffusion model for the primary supervision and proposes cascade DyNeRF to facilitate stable training. 
More recent work STAG4D~\cite{zeng2024stag4d} uses a multi-view diffusion model to initialize multi-view images anchored on input video frames and introduces a fusion strategy to improve the temporal consistency. 
However, the use of SDS loss in these works results in slow optimization speed and limits their applicability.
To improve the efficiency, Efficient4D \cite{pan2024fast} generates multi-view captures of dynamic 3D objects through SyncDreamer-T, while Diffusion$^2$\cite{yang2024diffusion} leverages orthogonal diffusion models to sample dense views. 
Despite the high efficiency, the assumptions of spatial-temporal conditional independence, multiple model utilization, and distinct-timestamp inference design inherently lead to 4D geometry inconsistencies.
Unlike previous methods, our framework is the first to integrate 4D spatial-temporal consistency into a singular video diffusion model.
% and it can obtain multi-timestamp cross-view supervision in one shot. 
More discussions about related works are in App.~\ref{app:related_work}.

% \vspace{-0.3cm}
\section{Method}
In this section, we provide a detailed illustration of the proposed framework, \textbf{Diffusion4D}, designed for the efficient and consistent generation of 4D content.
We initiate by formulating the problem and outlining the key objectives~\ref{sec:problem}. 
Then, we introduce our data curation strategy pivotal in acquiring a large-scale, high-quality 4D dataset~\ref{sec:data_curation}.
Subsequently, we delve into the methodology of applying the curated dataset to develop \textbf{4D-aware video diffusion models}, capable of synthesizing orbital views of dynamic 3D assets conditioning on various forms of prompts~\ref{sec:video_generation}.
Finally, we introduce the explicit construction of 4D assets using 4D models based on the Gaussian Splitting representations~\ref{sec:4d_recon}.

% \vspace{-0.2cm}
\subsection{Problem Setting and Key Objectives}
\label{sec:problem}
% Given a prompt as the condition, our primary objective is to generate spatially and temporally consistent multi-view images of 4D assets, followed by explicit 4D construction.
% % Inspired by the spatial-temporal consistency achieved in video diffusion models, our approach aims to migrate this property to the synthesis of orbital views around dynamic 3D assets.
% Our approach aims to migrate the temporal consistency embeded in video diffusion models to the spatial-temporal consistency in a 4D-aware video diffusion model.
Formally, given a prompt $y$, we aim to generate an orbital video $\mathcal{V}$ = ${\{I_i\in R^{H\times W\times 3}\}}_{i=1}^T$ around a dynamic 3D asset. This video comprises $T$ multi-view images captured at $T$ consecutive timestamps $\mathcal{T}$ = ${\{\tau_i}\}_{i=1}^T$ along a predefined camera pose trajectory, where $H$ and $W$ are the height and width dimensions of images. 
To simplify the problem, we put constraints on the camera pose trajectory, assuming that the camera always looks at the center of an object (origin of the world coordinates) with a fixed elevation angle and camera distance. Thus, the viewpoint can be uniquely specified by the azimuth angle. The azimuth angle uniformly increases from 0 to 360 degrees along $T$ timestamps, constructing a complete orbital video.
We aim to generate this orbital video with a 4D-aware video diffusion model that can iteratively denoise samples from the learned conditional distribution $p(\mathcal{V}|y)$, where $y$ can be text, single images, or static 3D content. 
At the end, explicit 4D construction is performed based on generated orbital video $\mathcal{V}$.

\subsection{Data Curation}
\label{sec:data_curation}

The development of the proposed 4D-aware video diffusion model requires a substantial amount of high-quality 4D assets. 
We curate a large-scale, high-quality dynamic 3D dataset sourced from the vast 3D data corpus of Objaverse-1.0 and Objaverse-XL~\cite{deitke2023objaverse,deitke2024objaverse}. 
Since the original Objaverse dataset primarily consists of static 3D assets and many of them are of low quality (e.g., partial scans, missing textures)~\cite{tang2024lgm}, we applied a series of empirical rules to filter the dataset. 
The curation process includes an initial selection with \textit{dynamic} label, removal of assets with subtle or overly dramatic motion, and out-of-boundary detection. 
In the initial step, we select 3D assets labeled as "dynamic". 
However, upon closer examination, we observe that many assets exhibit subtle or imperceptible movement, limiting their utility in learning dynamic properties in the proposed task. 
To mitigate this problem, we employ Structural Similarity Index Measure (SSIM) to evaluate the temporal dynamics of the assets. Specifically, for each 4D asset, we fix the camera pose at the front view and render three images at three distinct timestamps ($\tau_0, \tau_{T/4}$, and $\tau_{T/2}$), then compute two SSIM scores between them. 
Assets with both SSIM scores higher than a predetermined threshold $s_{high}=0.95$, suggesting high resemblance and little movement, are discarded.
We also observe that many samples exhibit significant geometric distortion or drastic appearance changes over time. 
Therefore, we empirically set an SSIM score of small value $s_{low}=0.4$ as the lower bound to filter out cases of poor quality.
Subsequently, for each remaining 4D asset, we render $T=24$ multi-view images following the camera pose trajectory and timeline settings as we defined in Sec. ~\ref{sec:problem}. 
In the final step, we handle cases where assets exhibit over-dramatic movement and extend out of the boundaries of the scene. We employ alpha maps to identify and remove such cases and ensure that only appropriately positioned dynamic assets are included in the curated dataset, enhancing the overall quality and coherence of the generated content.
This comprehensive strategy results in a total of 54K high-quality animated assets. 
We will release the IDs of the dynamic 3D assets in our curated dataset.

\begin{figure}
    \begin{center}
        \includegraphics[width=1.0\linewidth]
        {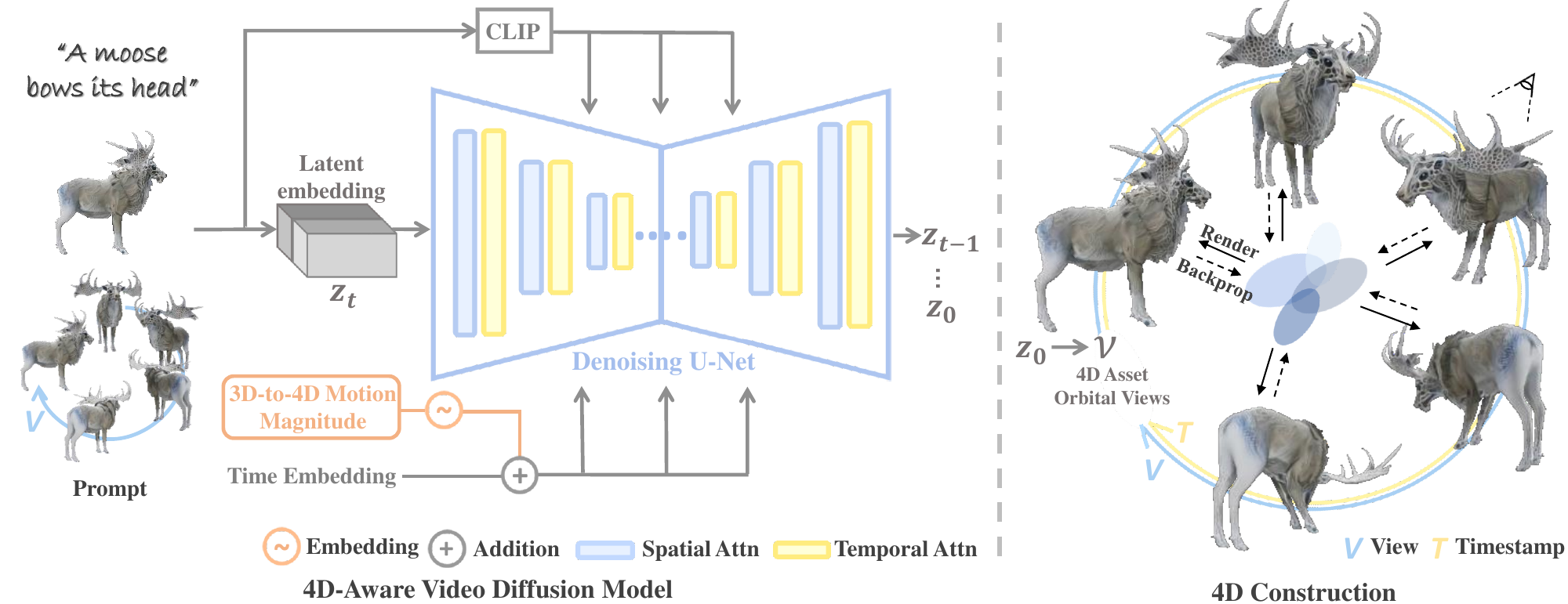}
        \caption{Our proposed Diffusion4D consists of a 4D-aware video diffusion model and explicit 4D construction, capable of synthesizing 4D assets conditioned on text, single images, or static 3D assets.}
        \label{fig:method}
    \end{center}
    % \vspace{-0.5cm}
\end{figure}

% Our method relies on a curated dataset of high-quality and large-scale dynamic 3D assets sourced from public benchmark 3D datasets. 

\subsection{4D-Aware Video Generation}
\label{sec:video_generation}

% Leveraging the self-curated dataset, we can obtain great amounts of orbital videos and finetune a large 3D-aware video diffusion model, to achieve the scalable generation of orbital videos for 4D assets with spatial-temporal consistency. 
% Leveraging the self-curated dataset, we can obtain great amounts of orbital videos to finetune the video diffusion models. Given that conventional video diffusion models designed for monocular video generation lack 3D geometry priors, we resort to the open-source 3D-aware video generative models, to achieve the scalable generation of orbital videos for 4D assets with spatial-temporal consistency. 
% By leveraging these models, we aim to achieve scalable generation of orbital videos for 4D assets with spatial-temporal consistency.
Utilizing the curated dataset, we can render a vast number of orbital videos of dynamic 3D assets to train a 4D-aware video diffusion model.
For pretrained model selection, given that conventional video diffusion models designed for monocular video generation lack 3D geometry priors, we resort to the recent works of 3D-aware video generation models~\cite{voleti2024sv3d,zuo2024videomv}. 
In the following, we first introduce the pretrained 3D-aware video diffusion model. Then, we describe how to adapt the model to our 4D-aware orbital video generation task. 
At this point, we focus on explicitly specifying the motion magnitude guidance, directly learning the 3D-to-4D dynamics in the training stage, and further augmenting the 3D object dynamics in the inference stage.
% Finally, we introduce how we utilize the curated dataset and customize the model architecture to accommodate various input prompt modalities.
Finally, we introduce how we utilize the curated dataset and customize the model architecture to accommodate various condition modalities.

\textbf{Pretrained 3D-aware video diffusion models.} 
The main idea of 3D-aware video diffusion models~\cite{voleti2024sv3d,zuo2024videomv} is to repurpose the temporal consistency in video generation models for the spatial 3D consistency of static 3D objects. 
% Taking advantage of conventional video diffusion models pretrained on large-scale high-quality natural video dataset, these methods finetune the models on large-scale 3D dataset to generate multi-view videos around the static 3D assets.
% These models typically capitalize on the conventional video diffusion models pretrained on extensive natural videos. 
Capitalizing on pretrained video diffusion models~\cite{blattmann2023stable,wang2023modelscope,zhang2023i2vgen}, these models are finetuned on large-scale datasets of high-quality 3D assets. It facilitates them to generate smooth and consistent orbital views of static 3D objects with user-specified camera pose conditions.
In our method, we inherit the 3D geometry consistency in these models and extend it to modeling the spatial-temporal consistency in orbital views of 4D objects.

\textbf{Vanilla 4D-aware video diffusion models.}
Modern video diffusion models typically carry out diffusion and denoising processes in latent space~\cite{blattmann2023align, yu2023video}.
In our specific task, given an orbital video $\mathcal{V}$ around a 4D asset, which is rendered from curated dataset following the camera position trajectory introduced in Sec.~\ref{sec:problem}, we first use a pretrained encoder to encode images into a compressed latent $z_0\in R^{T\times h\times w\times c}$. 
The $h,w,c$ respectively denote the height, width, and channel dimension of the latent representation.
The diffusion forward process samples a time step $t$ and adds noise $\epsilon_t$ to source input and obtain $z_t$.
A denoising network $\epsilon_{\theta}$, parameterized by $\theta$, is trained to predict the added noise conditioned on $y$, with a noise prediction loss
$\mathcal{L}_{ldm}  = {\left\|\epsilon_t-\epsilon_{\theta}(z_{t},y,t)\right\|}_{2}^{2}$.
In the inference stage, as shown in Fig.~\ref{fig:method}, given an initial random gaussian noise and prompt condition $y$, the denoising network predicts the added noise to iteratively denoise the latent embedding. 
A denoised latent $\hat{z}_{0}$ is finally obtained, which is decoded via a pretrained decoder to recover a high-fidelity orbital video.

% In the inference stage, classifier-free guidance is usually introduced to the denoising sampling loop. 
% Given an initial Gaussian noise, the denoising network $\epsilon_{\theta}$ predicts the added noise iteratively, which is formulated as:
% \begin{equation}
% \label{eq:cfg}
% \overline{\epsilon}_{\theta}(z_t,y,t) = \epsilon_{\theta}(z_t, y, t) + w(\epsilon_{\theta}(z_t,y,t)-\epsilon_{\theta}(z_t,t))
% \end{equation},
% where $\epsilon_{\theta}(z_t,t)$ is unconditional noise prediction and $w$ is the classifier-free guidance scaler. 
% A denoised latent $z_{0}$ is finally obtained, which is decoded via a pretrained decoder to recover a high-fidelity video.

\textbf{3D-to-4D motion magnitude guidance and reconstruction}.
For the 4D-aware video generation, one of the most interesting attributes is the dynamic strength of the 3D assets. 
In an effort to enhance the control over the 3D-to-4D dynamic strength, we introduce the motion magnitude guidance into the diffusion model.
To begin with, we need to quantitatively measure the motion magnitude of the moving objects.
In contrast to the previous video models~\cite{dai2023fine} that measure the motion magnitude by computing the inter-frame differences, the appearance differences across frames in $\mathcal{V}$ originate from both camera-view changes and object movement.
To remove the influence from the camera-view changes, for each dynamic 3D object, we also render an orbital video $\bar{\mathcal{V}}$ = ${\{\bar{I}_i\in R^{H\times W\times 3}\}}_{i=1}^T$ around the static 3D asset, consisting of $T$ multi-view images captured at the same camera poses as $\mathcal{V}$ at timestamp $\tau_0$.
Consecutively, we propose a metric named \textbf{3D-to-4D motion magnitude} \textbf{${m}$} measuring the dynamic strength:

\begin{equation}
m(\mathcal{V}) = \frac{1}{T}\sum_{i=1}^T || I_i - \bar{I}_i ||_{2}^{2}.
\end{equation}

Motivated by previous works~\cite{zuo2024videomv}, we use a two-layer multi-layer perception (MLP) to extract a motion magnitude embedding, which is combined with the time embedding and injected into the denosing network. The noise prediction function by the denoising network accordingly changes to $\epsilon_{\theta}(z_{t},y,m,t)$. This design allows the user to explicitly control the motion magnitude in inference stage. The impact of the motion magnitude guidance on the generation results is illustrated in Fig.~\ref{fig:ablation}.

% During model training, to further encourage the denoising network to learn the dynamics of the 3D objects, we introduce the motion magnitude reconstruction loss to enable direct learning of the dynamics strength. 
% The intuition comes from explicitly supervising the reconstruction of the 3D to 4D dynamics.
In the training phase, to encourage the denoising network to learn the 3D-to-4D dynamics, we incorporate the motion magnitude reconstruction loss. 
This loss facilitates the direct learning of dynamic strength with explicit supervision over motion magnitude from 3D to 4D dynamics. 
For the sake of computation cost, we apply the reconstruction loss in the latent space and it is formulated as:
\begin{equation}
\mathcal{L}_{mr} = || m(z_0) -  m(\hat{z}_0) ||_{2}^{2},~ m(z_0) = \frac{1}{T} || z_0 - \bar{z}_0 ||_{2}^{2}.
\end{equation},
where $\hat{z}_0$ denotes the estimated clean video latent during training, the $\bar{z}_0$ denotes the compressed latent of $\bar{\mathcal{V}}$ by the pretrained encoder. Our diffusion model training loss is formulated by combining latent diffusion loss and motion magnitude reconstruction loss with a weight $\omega$:
$\mathcal{L} = \mathcal{L}_{ldm} + \omega\mathcal{L}_{mr}$.

\textbf{3D-aware classifier-free guidance}.
% During the inference phase, we notice that there are rare circumstances where objects exhibit minimal motion. 
% The generated videos only capture the view changes and show orbital views of a nearly static 3D asset. 
To further augment the dynamic strength of 3D objects, we draw inspiration from classifier-free guidance~\cite{ho2022classifier} and propose a 3D-aware classifier-free guidance.
It uses the pretrained 3D-aware video diffusion model, formulated as $\bar{\epsilon}_{\theta}$, to provide classifier-free guidance during the inference stage. 
Combining the prompt condition $y$, motion magnitude guidance $m$, the unconditional noise prediction $\epsilon_{\theta}(z_t,t)$ and 3D-aware noise prediction $\bar{\epsilon}_{\theta}(z_t,y,t)$, the denoising step is reformulated as:
\begin{equation}
\hat{\epsilon}_{\theta}(z_t, y, m, t) = \epsilon_{\theta}(z_t,y,m,t) + w_1(\epsilon_{\theta}(z_t,y,m,t)-\epsilon_{\theta}(z_t,t)) + w_2(\epsilon_{\theta}(z_t,y,m,t)-\bar{\epsilon}_{\theta}(z_t,y,t))
\end{equation},
where $\hat{\epsilon}_{\theta}(z_t, y, m, t)$ is joint noise prediction, and $w_1$ and $w_2$ are the classfier-free guidance scalers.
% The generated video offers comprehensive information for the downstream explicit modeling with 4D representations.

\textbf{Generation with various condition modalities.} 
In the above formulations, we use $y$ as a general prompt condition.
Thanks to the versatility of our high-quality 4D dataset and the flexibility of the 3D-aware video diffusion model architecture, our framework can readily accommodate diverse prompt modalities, including text, single images, or static 3D contents. 
\textbf{For the text condition}, we obtain the text description of each dynamic 3D asset from the prior work~\cite{luo2024scalable}, and the text embedding extracted by CLIP model is fed into the diffusion model via the cross-attention mechanism. You can refer to~\cite{wang2023modelscope} for more details.
\textbf{For the image condition}, we use the first view image $I_0$ in $\mathcal{V}$ captured at timestamp $\tau_0$ as the reference image. The image condition is injected into the diffusion model with both cross-attention mechanism and feature concatenation. Please refer to~\cite{zhang2023i2vgen, blattmann2023stable} for more details.
\textbf{For static 3D content condition}, we use $\bar{\mathcal{V}}$ as the reference video. Similar to the image condition, the video features are extracted by pretrained encoder and and fed into the diffusion model via feature concatenation.
This versatility allows our framework to effectively respond to different condition modalities, facilitating seamless integration into a wide range of applications and scenarios.

\vspace{-0.3cm}
\subsection{Coarse-to-Fine 4D Construction}
\label{sec:4d_recon}

The spatial-temporal consistent multi-view videos generated by our 4D-aware video diffusion model offer comprehensive information about the 3D geometry and motions of dynamic 3D assets.
At this stage, we explicitly model the 4D assets with 4D Gaussian Splatting (4DGS)~\cite{ren2023dreamgaussian4d} owing to its explicit representation, powerful fitting capabilities, and efficient optimization with dense-view supervision.
% Each Gaussian is defined with center position $\mu \in \mathbb{R}^3$, covariance matrix $\Sigma\in \mathbb{R}^{3\times3}$, color $c \in \mathbb{R}^3$ and opacity $\alpha \in \mathbb{R}^1$. It can be formulated as:
% \begin{equation}
% G(x)=e^{-\frac{1}{2}{(x)}^{T}\Sigma^{-1}(x)},
% \label{eq:3dGS}
% \end{equation}
% 4DGS employs an efficient deformation field network to accurately model Gaussian motions and shape deformations, where adjacent Gaussians are connected via a hexplane(cite hexplane kplane) to keep spatial-temporal consistency.%加上mlp的公式
% When we directly train the 4DGS on the generated multi-view video ${\mathcal{V}}$, we find that the model fails to construct a decent 3D geometry due to the lack of supervision signals.
When we directly train the 4DGS on the generated multi-view video ${\mathcal{V}}$, we find that the model does not perform well in capturing the 3D geometry details.
The visualization can be found in Fig.~\ref{fig:ablation}(e).
Therefore, we augment the supervision signals and propose a coarse-to-fine construction strategy. 
In the coarse stage, given the first-view image ${I_0}$ in ${\mathcal{V}}$, we use the pretrained 3D-aware video diffusion model to generate an orbital-view video ${\bar{\mathcal{V}}}^{'}$ of the static 3D object.
As our 4D-aware video diffusion model is finetuned on this model, we observe high 3D geometry consistency between ${\bar{\mathcal{V}}}^{'}$ and ${\mathcal{V}}$.
We merge them together to train the 4DGS in the coarse training stage. 
Actually, for static 3D content conditioned generation, we can readily replace ${\bar{\mathcal{V}}}^{'}$ with conditional signal $\bar{\mathcal{V}}$ into the image set for coarse training.
In the fine training stage, we use ${\mathcal{V}}$ only to further tune the 4DGS to improve the spatial-temporal consistency.
Thanks to the 4D consistency in our generated videos, we can achieve precise pixel-level matching across different views and timestamps. We follow~\cite{wu20234d} and use $L_1$ and $L_{lpips}$~\cite{zhang2018unreasonable} losses for optimization. 
To enforce geometry smoothness, we also involve depth smoothness loss as regularizer~\cite{charbonnier1994two,niemeyer2022regnerf}.
The total loss is formulated as 
% \begin{equation}
$\mathcal{L}_{gs} = \lambda_{l1}\mathcal{L}_{1} + \lambda_{lpips}\mathcal{L}_{lpips}+\lambda_{depth}\mathcal{L}_{depth}$,
% \end{equation},
where $\lambda_{l1}$, $\lambda_{lpips}$ and $\lambda_{depth}$ are losses weights.

% We first use our video diffusion outputs to reconstruct 4D sence. The video outputs are 24 frame monocular video. We use $L_1$ and $L_{lpips}$ loss to supervise gaussians. Due to monocular video reconstruction misses geometry and detail information in novel view, only using dynamic video outputs fail to reconstruct a accurate result. To augument training data for 4DGS reconstruction, we feed first refernce image into origin VideoMV to render static multi views frames. Then we use static and dynamic multi views frames to optimize 4DGS. As a result, we improve the reconstruct quality and robustness at novel views, without additional modules and computation.

%%%%%%%%%%%%%%%%%%%%%%%%%%%%%%%%%%%%%%%%%%%%%%%%%%%%%%%%%%%%

\begin{figure}
    \begin{center}
        \includegraphics[width=1.0\linewidth]
        {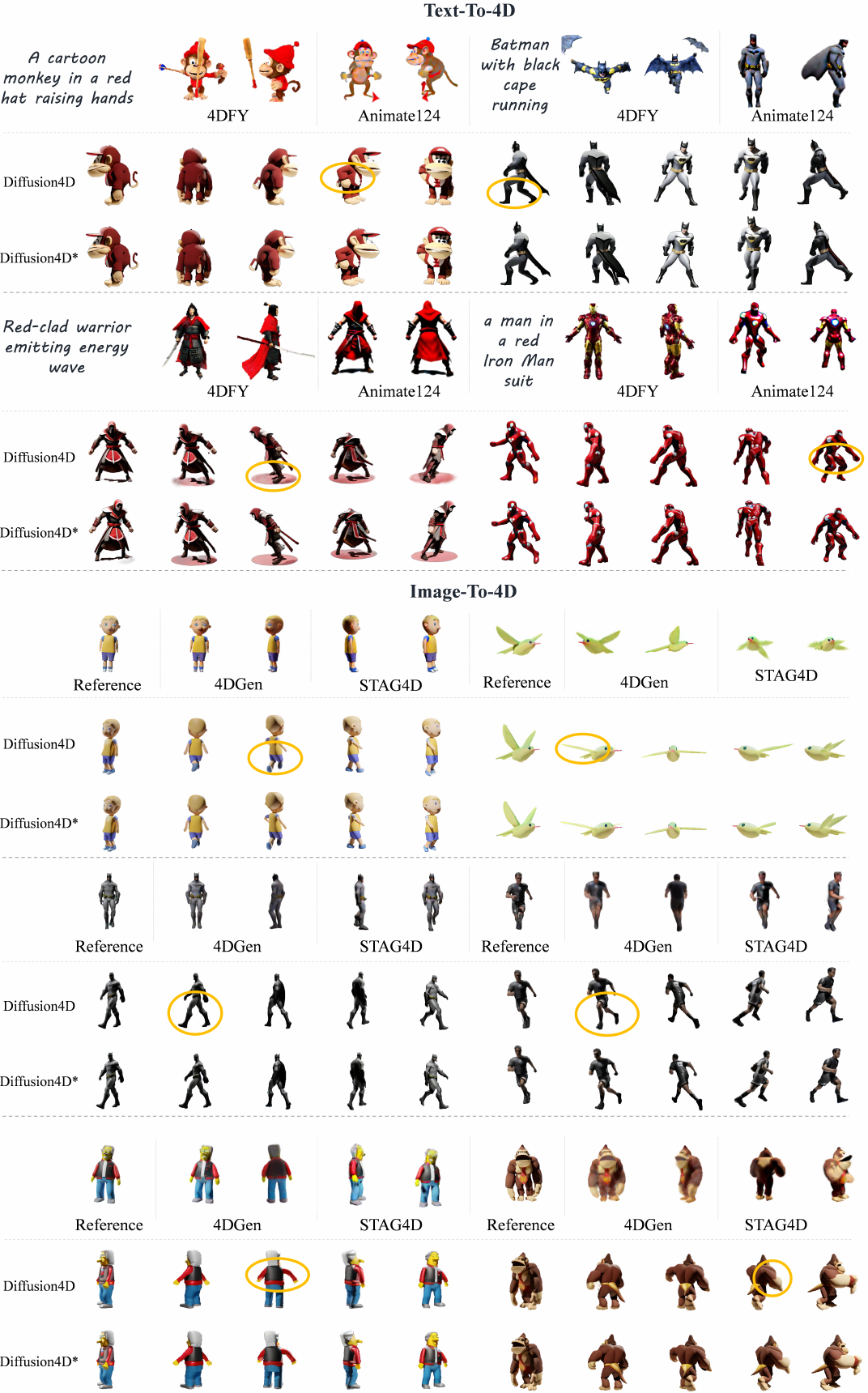}
        % \vspace{-0.2cm}
        \caption{Qualitative comparisons between Diffusion4D and other baselines in Text-to-4D (upper) and Image-to-4D (lower) generation. For our method, we show five views from consecutive timestamps. (* results from 4D-aware video diffusion model).}
        \label{fig:main_result1}
    \end{center}
    % \vspace{-0.3cm}
\end{figure}

\begin{figure}
    \begin{center}
        \includegraphics[width=1.0\linewidth]
        {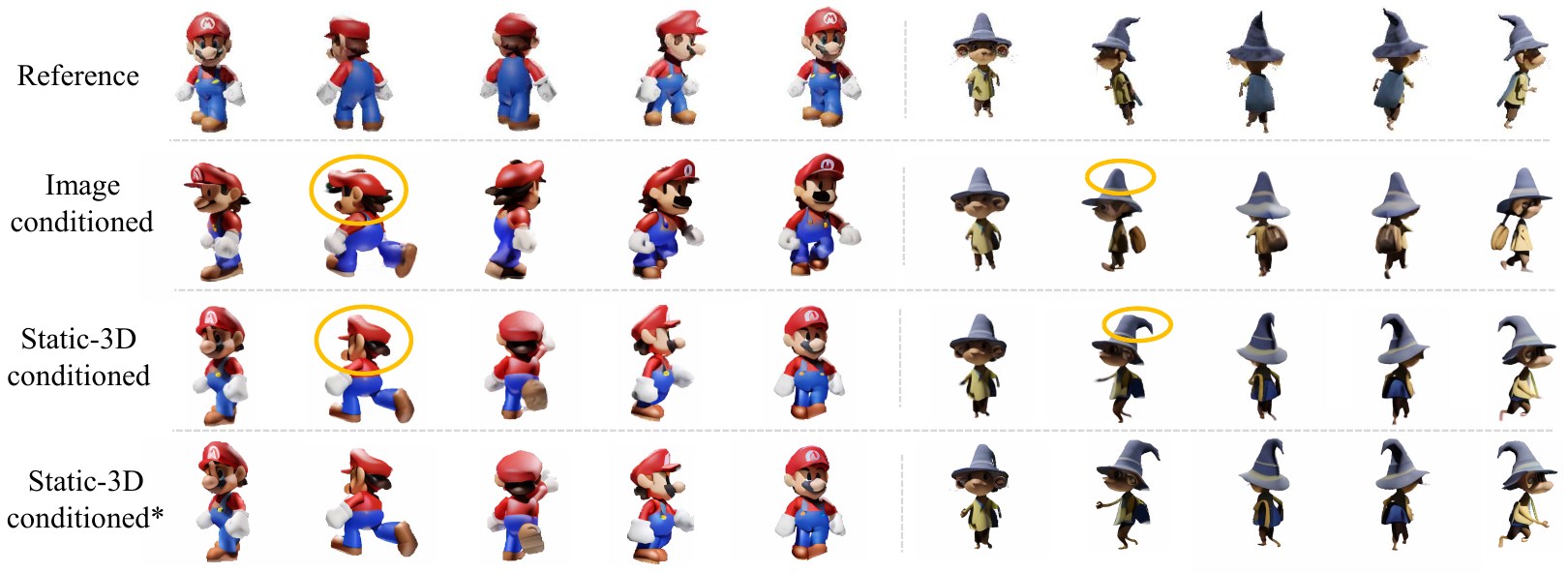}
        \caption{Visualization of Static-3D conditioned Diffusion4D. The first row shows the conditions, and the rest shows the results. (* results from 4D-aware video diffusion model.)}
        \label{fig:3d}
    \end{center}
    \vspace{-0.2cm}
\end{figure}

\vspace{-0.2cm}
\section{Experimentation}
\subsection{Experiment Setup}
\label{sec:exp_setup}
We employ Diffusion4D to generate 4D assets conditioned on multiple modalities of prompts, i.e. text, single image, and static 3D content.
Leveraging the curated dataset, for each dynamic 3D asset, we render two 360-degree orbital videos $\mathcal{V}'$s with $T=24$, one starting at the front view and the other starting at a random azimuth angle. 
We fix the elevation angle as 0 and the camera distance as 2-meter. The image resolution is chosen at $256\times 256$ across the experiments. 
In preparation for static 3D content-conditioned generation, following the camera poses of each rendered video $\mathcal{V}$, we also render two 360-degree orbital video $\bar{\mathcal{V}}'$s of each static 3D asset fixed at initial timestamp $\tau_{0}$. 
We use VideoMV~\cite{zuo2024videomv} as the pretrained 3D-aware video diffusion model, which is trained on a large scale of static 3D assets of superior quality.
Specifically, for text-conditioned 4D-aware video generation, we adopt the text-to-3D video generator~\cite{zuo2024videomv} that is based on the ModelScopeT2V~\cite{wang2023modelscope} model architecture. 
For image and video conditions, we adopt the image-to-3D video generator~\cite{zuo2024videomv} that is finetuned from I2VGen-XL~\cite{zhang2023i2vgen}.
We train our 4D-aware video diffusion model for 4k iterations with a constant learning rate of $3\times 10^{-5}$.
% The motion magnitude reconstruction loss weight $\omega$ is set to be $5\times 10^{-4}$.
During the sampling stage, we use text, front-view images, front-view started orbital videos of static 3D assets as conditions.
we use DDIM~\cite{song2020denoising} sampling with sampling step 50, and  $w_1=7.0$ and $w_2=0.5$ in classifier-free guidance. 
In the 4D construction stage, we optimized 4DGS representation for 5,000 iterations in the coarse stage and 2,000 iterations in the fine stage.
We provide more experimental details in App.~\ref{app:exp_detail}.

% \vspace{-0.1cm}
\subsection{Metric}
\label{sec:metric}
From the curated dataset, we leave out 20 cases as the test set for evaluation. 
For quantitative assessment, we first use the CLIP~\cite{radford2021learning} score to evaluate the semantic correlation between the prompts (reference) and synthesized images (target).
In each case, depending on the condition modality, the text description or front-view image serves as the reference.
For the diffusion generation, we use all 24 images in orbital videos as targets, and for 4D construction, we uniformly render 36 orbital views from constructed 4D assets as targets (CLIP-O). 
We also measure using only front-view images as targets(CLIP-F).
To evaluate the appearance and texture quality and the spatial-temporal consistency of the synthesized images, we also use the following metrics, i.e. LPIPS~\cite{zhang2018unreasonable}, PSNR, SSIM~\cite{wang2004image}, and FVD~\cite{unterthiner2019fvd}, to help assess image- and static 3D content-conditioned generation. 
For these evaluations, we use ground truth images rendered from the dynamic 3D assets as references. 
Images generated by the diffusion model or rendered from constructed 4D assets are used as targets.
Metrics are computed between ground truth images and synthesized images from the same camera pose.
The same procedure is applied to calculate the scores for baseline methods.
We also conduct qualitative comparisons through a user study involving 30 human evaluators from diverse backgrounds.
Participants are asked to specify their preference for rendered orbital videos around 4D assets based on four properties, following the approach in \cite{bahmani20234d}: 3D geometry consistency (3DC), appearance quality(AQ), motion fidelity(MF), and text alignment(TA). 
We report the percentage of users who prefer each method overall and based on each of the four properties.

\subsection{Main Results}
\label{sef:mainresults}
\textbf{Text-conditioned and image-conditioned 4D generation.}
We take 4DFY~\cite{bahmani20234d} and Animate124~\cite{zhao2023animate124} as baselines for text-conditioned generation, and 4DGen\cite{yin20234dgen} and STAG4D\cite{zeng2024stag4d} as baselines for image-conditioned generation, due to their remarkable performance and adaptability to the task settings.
% For text-conditioned generation, we compare our method with the baselines of 4DFY~\cite{bahmani20234d} and Animate124~\cite{zhao2023animate124}, and for image-conditioned generation, we compare with 4DGen\cite{yin20234dgen} and STAG4D\cite{zeng2024stag4d}  due to their remarkable performance and adaptability to our task setting.
In Fig.~\ref{fig:main_result1} we provide a detailed comparison between our method and baselines with various prompts.
For the baselines, we provide two views at starting and ending timestamps. 
For our method, we visualize the spatial-temporal renderings from 4D constructions at five consecutive timestamps in multiple views.
We also show the generated multi-view images from our video diffusion models (denoted with *).
Our efficient and elegant pipeline is capable of generating diverse and realistic 4D assets with satisfactory geometrical and temporal consistency.
While the baselines also synthesize 4D assets, their results exhibit very limited or even invisible motion. 
In contrast, our results show apparent motions of great fidelity. 
As highlighted by the contours in Fig.~\ref{fig:main_result1}, our animations include cartoon characters and humans stepping forward, running or raising arms, birds flapping wings, and lights changing.
Our method also provides more detailed appearances.
The quantitative results in Tab.~\ref{tab:results} demonstrate that our method outperforms previous approaches across all metrics and user preferences. Compared with the state-of-the-art SDS-based methods, which involve sophisticated and time-consuming optimization, our method is much more efficient and produces more favorable results. Users showed a strong preference for Diffusion4D over other baselines, especially in 3D geometry consistency and motion fidelity. Also, as 4D construction is based on the images from the diffusion models, we can observe that diffusion outputs perform slightly better than rendered outputs.

\begin{table}
\centering
\scriptsize
    \caption{Quantitative comparison between our method with other baselines under different task settings. The human study includes 3D geometry consistency(3DC), appearance quality(AQ), motion fidelity(MF), text alignment(TA), and overall score. By default, Diffusion4D suggests results from 4D construction, and Diffusion4D* suggests results from our diffusion models. } 
    \vspace{-0.3cm}
    \label{tab:results}
    \begin{center}
    \resizebox{\columnwidth}{!}{
    \begin{tabular}{lccccccc|ccccc}
        \toprule
         & \multicolumn{7}{c}{\textit{Metrics}} & \multicolumn{5}{c}{\textit{Human Preference}}\\
        \textit{Method} & CLIP-F$\uparrow$ & CLIP-O$\uparrow$ & Generation time$\downarrow$ & SSIM$\uparrow$ & PSNR$\uparrow$ & LPIPS$\downarrow$  & FVD$\downarrow$  &  3DC & AQ & MF & TA & Overall \\
        \midrule
               % \midrule
        \textit{Text-to-4D} & \multicolumn{12}{c}{}  \\
        \midrule
        4DFY~\cite{bahmani20234d} & 0.78 & 0.61 & 23h &\multicolumn{4}{c|}{---} & 26\% & 34\% & 25\% & 37\% & 29\%\\
        Animate124~\cite{zhao2023animate124} & 0.75 & 0.58  & 9h &\multicolumn{4}{c|}{---} & 22\% & 28\% & 19\% & 23\% & 22\%\\
        Diffusion4D & \textbf{0.81}&\textbf{0.65}  & \textbf{8m}  &\multicolumn{4}{c|}{---} & \textbf{52\%} & \textbf{38\%} & \textbf{56\%} &\textbf{40\%} & \textbf{49\%} \\
        Diffusion4D$^{\ast}$ & \textbf{0.82} & \textbf{0.69} &--  &\multicolumn{4}{c|}{---}& \multicolumn{5}{c}{---} \\
        
        \midrule
        \textit{Image-to-4D} & \multicolumn{12}{c}{}  \\
        \midrule
        4DGen~\cite{yin20234dgen} & 0.84 & 0.71 & 1h30m & 0.69 & 14.4 & 0.31 & 736.6       & 18\% & 22\% & 18\% & 29\% & 22\%\\
        STAG4D~\cite{zeng2024stag4d} & 0.86 & 0.72 & 2h30m & 0.76 & 15.2 & 0.27 & 675.4 & 15\% & 25\% & 26\% & 33\% & 24\%\\
        Diffusion4D & \textbf{0.89} & \textbf{0.75} & \textbf{8m} & \textbf{0.83} & \textbf{16.7} & \textbf{0.21} & \textbf{560.8}                     & \textbf{67\%} & \textbf{53\%} & \textbf{56\%} & \textbf{38\%} & \textbf{54\%} \\
        Diffusion4D$^{\ast}$ & \textbf{0.90} & \textbf{0.80} & -- & 0.82 & \textbf{16.8} & \textbf{0.19} & \textbf{490.2} & \multicolumn{5}{c}{---} \\
        
        \midrule
        \textit{Static 3D content-to-4D} & \multicolumn{12}{c}{}  \\\midrule
        Diffusion4D & 0.88 & 0.77 & 8m & 0.82 & 16.8 & 0.19 & 544.7 & \multicolumn{5}{c}{---} \\
        Diffusion4D$^{\ast}$ & \textbf{0.91} & \textbf{0.81} & -- & \textbf{0.83} & \textbf{17.2} & \textbf{0.18} & \textbf{482.4} & \multicolumn{5}{c}{---} \\
        \bottomrule
    \end{tabular}}
    \end{center}
    % \vskip -0.15in
\end{table}

\begin{figure}
    \begin{center}
        \includegraphics[width=1.0\linewidth]
        {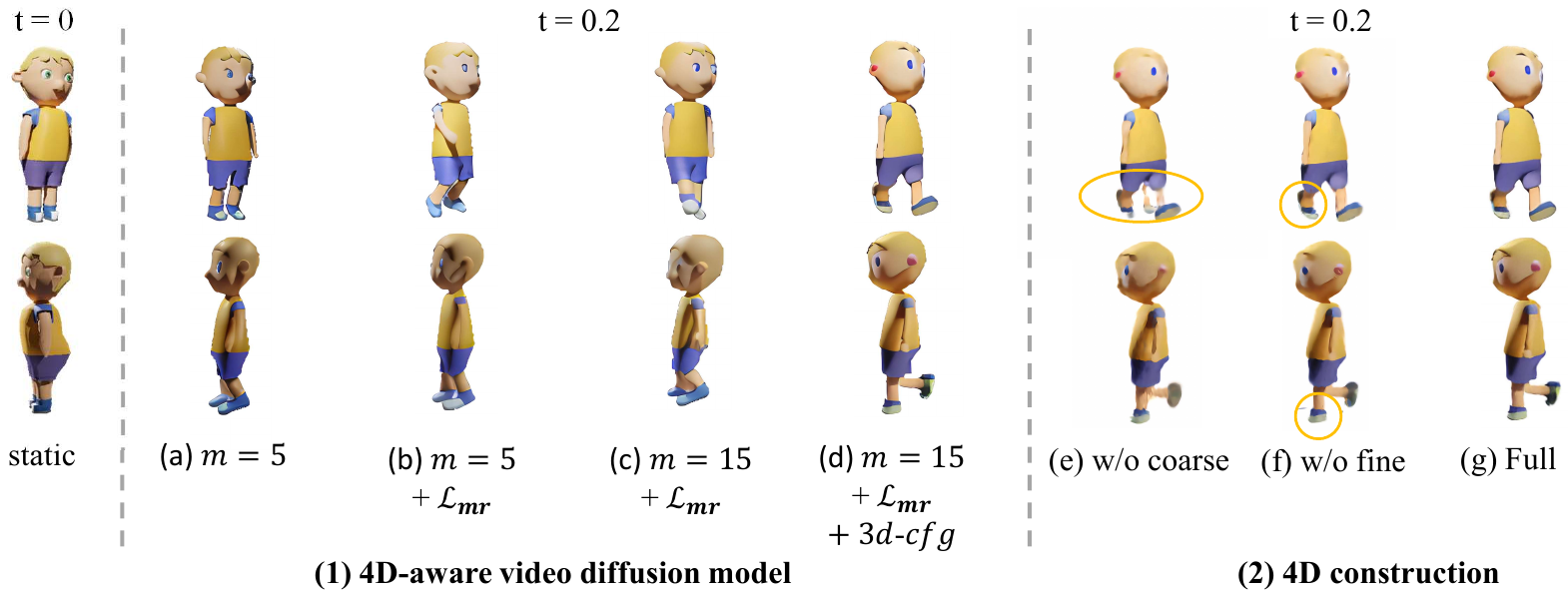}
        \caption{Ablation study of different components of our proposed framework.}
        \label{fig:ablation}
    \end{center}
    \vspace{-0.3cm}
\end{figure}

\textbf{Static 3D content-conditioned 4D generation.}
We visualized the results of static 3D content-conditioned generation in Fig.~\ref{fig:3d}. 
Our framework successfully activates static 3D assets and generates spatial-temporal consistent 4D assets.
Due to the lack of direct baselines, we compare our image-conditioned 4D generation with 3D content-conditioned generation.
The difference between these two settings is that the former relies only on a single image, whereas the latter has access to dense views of the 3D assets.
While both can generate realistic dynamic 3D assets, the latter can generate more 3D geometry-consistent objects.
As shown in the last two rows in Fig.~\ref{fig:3d}, Mario and anthropomorphic mouse closely follows the appearance and geometry of the prompts, particularly from the side and back views.
The quantitative results in Tab.~\ref{tab:results} indicate that the static 3D content-conditioned setting achieves the best performance among all tasks, thanks to the access to dense views of ssource assets. 
%\textbf{We also provide videos in Supp. for better demonstration.}
% \vspace{-0.1cm}
\subsection{Ablation Study}

\begin{table}%{r}{8.5cm}
	\centering
       % \vspace{-0.5cm}
        \caption{Ablation study on the effect of proposed components on 4D construction. * denotes results from 4D-aware video diffusion models. Those without * denote results from 4D construction.} 
        \label{tab:ablation}
        \resizebox{0.6\columnwidth}{!}{
        \begin{tabular}{lcccccc}
        \toprule
         
        \textit{\textit{Image-to-4D}} & CLIP-F$\uparrow$ & CLIP-O$\uparrow$ & SSIM$\uparrow$ & PSNR$\uparrow$  & LPIPS$\downarrow$ & FVD$\downarrow$  \\
        
        \midrule
        w/o $\mathcal{L}_{mr}$ & 0.84 & 0.73 & 0.78 & 15.4 & 0.26 & 602.5 \\
        w/o $\mathcal{L}_{mr}$$^{\ast}$ & 0.86 & 0.77 & 0.79 & 15.8 & 0.23 & 524.6 \\
        w/o coarse stage & 0.79 & 0.70 & 0.72  & 14.5 & 0.32 & 651.2 \\
        w/o fine stage & 0.86 & 0.74 & 0.77 & 15.2 & 0.25 & 581.4 \\
        Full model & \textbf{0.89} & \textbf{0.75} & \textbf{0.83} & \textbf{16.7} & \textbf{0.21} & \textbf{560.8} \\
        Full model$^{\ast}$ & \textbf{0.90} & \textbf{0.80} & \textbf{0.82}& \textbf{16.8} & \textbf{0.19} & \textbf{490.2} \\
        \bottomrule
         % \vspace{-0.6cm}
    \end{tabular}}
\end{table}

We conduct an analysis on the effect of various design components in our framework with the typical image-to-4D generation task, and the results are shown in Fig.~\ref{fig:ablation} and Tab.~\ref{tab:ablation}.
In the 4D-aware diffusion model, we incorporated multiple features to enhance the dynamics of 3D assets. We added each feature incrementally to demonstrate their impact.
In Fig.~\ref{fig:ablation} (1), we can observe that training without motion magnitude reconstruction loss results in nearly invisible movement (a), while incorporating it introduces subtle motion (b). 
Increasing the motion magnitude guidance ($m$) augments the dynamics of the kid (c). 
The involvement of 3D-aware classifier-free guidance significantly augments the dynamics of the kid (d). 
In the 4D construction stage, as you can observe in Fig.~\ref{fig:ablation} (2), training without the coarse stage leads to incomplete 3D geometry (e). Comparing (f) and (g), we can observe that adding the fine stage enhances the appearance and texture details of the generated 4D assets. 
Results in Tab.~\ref{tab:ablation} indicate that the removal of each proposed component leads to a noticeable drop in performance metrics.
In summary, the full model delivers the best results both quantitatively and qualitatively.

% \section{Limitations and Future work}
% \vspace{-0.1cm}
\section{Conclusion}
In this work, we introduced Diffusion4D, an efficient and scalable framework for spatial-temporal consistent 4D generation. 
Motivated by the prior explorations using video diffusion model to generate orbital views of static 3D assets, we migrate the temporal consistency in the video diffusion model to spatial-temporal consistency in 4D generation. 
Leveraging a meticulously curated dynamic 3D dataset, we developed a 4D-aware video diffusion model capable of synthesizing orbital views of dynamic 3D assets. 
We incorporate 3D-to-4D motion magnitude guidance and the motion magnitude reconstruction loss to enhance the dynamics learning and control. 3D-aware classifier-free guidance is introduced to further augment the dynamic strength. We fulfill explicit 4D construction with a coarse-to-fine learning strategy.
Extensive experiments demonstrated that our method surpasses prior state-of-the-art techniques in terms of generation efficiency and 4D geometry consistency across various prompt modalities. This innovative approach not only addresses the limitations of previous methods but also sets a new benchmark for 4D content generation.

\clearpage

\bibliographystyle{plain}
\bibliography{neurips_2024}

\begin{thebibliography}{10}

\bibitem{bahmani20234d}
Sherwin Bahmani, Ivan Skorokhodov, Victor Rong, Gordon Wetzstein, Leonidas Guibas, Peter Wonka, Sergey Tulyakov, Jeong~Joon Park, Andrea Tagliasacchi, and David~B Lindell.
\newblock 4d-fy: Text-to-4d generation using hybrid score distillation sampling.
\newblock {\em arXiv preprint arXiv:2311.17984}, 2023.

\bibitem{blattmann2023stable}
Andreas Blattmann, Tim Dockhorn, Sumith Kulal, Daniel Mendelevitch, Maciej Kilian, Dominik Lorenz, Yam Levi, Zion English, Vikram Voleti, Adam Letts, et~al.
\newblock Stable video diffusion: Scaling latent video diffusion models to large datasets.
\newblock {\em arXiv preprint arXiv:2311.15127}, 2023.

\bibitem{blattmann2023align}
Andreas Blattmann, Robin Rombach, Huan Ling, Tim Dockhorn, Seung~Wook Kim, Sanja Fidler, and Karsten Kreis.
\newblock Align your latents: High-resolution video synthesis with latent diffusion models.
\newblock In {\em Proceedings of the IEEE/CVF Conference on Computer Vision and Pattern Recognition}, pages 22563--22575, 2023.

\bibitem{charbonnier1994two}
Pierre Charbonnier, Laure Blanc-Feraud, Gilles Aubert, and Michel Barlaud.
\newblock Two deterministic half-quadratic regularization algorithms for computed imaging.
\newblock In {\em Proceedings of 1st international conference on image processing}, volume~2, pages 168--172. IEEE, 1994.

\bibitem{chu2024dreamscene4d}
Wen-Hsuan Chu, Lei Ke, and Katerina Fragkiadaki.
\newblock Dreamscene4d: Dynamic multi-object scene generation from monocular videos.
\newblock {\em arXiv preprint arXiv:2405.02280}, 2024.

\bibitem{dai2023fine}
Zuozhuo Dai, Zhenghao Zhang, Yao Yao, Bingxue Qiu, Siyu Zhu, Long Qin, and Weizhi Wang.
\newblock Fine-grained open domain image animation with motion guidance.
\newblock {\em arXiv preprint arXiv:2311.12886}, 2023.

\bibitem{deitke2024objaverse}
Matt Deitke, Ruoshi Liu, Matthew Wallingford, Huong Ngo, Oscar Michel, Aditya Kusupati, Alan Fan, Christian Laforte, Vikram Voleti, Samir~Yitzhak Gadre, et~al.
\newblock Objaverse-xl: A universe of 10m+ 3d objects.
\newblock {\em Advances in Neural Information Processing Systems}, 36, 2024.

\bibitem{deitke2023objaverse}
Matt Deitke, Dustin Schwenk, Jordi Salvador, Luca Weihs, Oscar Michel, Eli VanderBilt, Ludwig Schmidt, Kiana Ehsani, Aniruddha Kembhavi, and Ali Farhadi.
\newblock Objaverse: A universe of annotated 3d objects.
\newblock In {\em Proceedings of the IEEE/CVF Conference on Computer Vision and Pattern Recognition}, pages 13142--13153, 2023.

\bibitem{gao2024gaussianflow}
Quankai Gao, Qiangeng Xu, Zhe Cao, Ben Mildenhall, Wenchao Ma, Le~Chen, Danhang Tang, and Ulrich Neumann.
\newblock Gaussianflow: Splatting gaussian dynamics for 4d content creation.
\newblock {\em arXiv preprint arXiv:2403.12365}, 2024.

\bibitem{gupta2023photorealistic}
Agrim Gupta, Lijun Yu, Kihyuk Sohn, Xiuye Gu, Meera Hahn, Li~Fei-Fei, Irfan Essa, Lu~Jiang, and Jos{\'e} Lezama.
\newblock Photorealistic video generation with diffusion models.
\newblock {\em arXiv preprint arXiv:2312.06662}, 2023.

\bibitem{ho2022imagen}
Jonathan Ho, William Chan, Chitwan Saharia, Jay Whang, Ruiqi Gao, Alexey Gritsenko, Diederik~P Kingma, Ben Poole, Mohammad Norouzi, David~J Fleet, et~al.
\newblock Imagen video: High definition video generation with diffusion models.
\newblock {\em arXiv preprint arXiv:2210.02303}, 2022.

\bibitem{ho2022classifier}
Jonathan Ho and Tim Salimans.
\newblock Classifier-free diffusion guidance.
\newblock {\em arXiv preprint arXiv:2207.12598}, 2022.

\bibitem{jiang2023consistent4d}
Yanqin Jiang, Li~Zhang, Jin Gao, Weimin Hu, and Yao Yao.
\newblock Consistent4d: Consistent 360 $\{$$\backslash$deg$\}$ dynamic object generation from monocular video.
\newblock {\em arXiv preprint arXiv:2311.02848}, 2023.

\bibitem{jun2023shap}
Heewoo Jun and Alex Nichol.
\newblock Shap-e: Generating conditional 3d implicit functions.
\newblock {\em arXiv preprint arXiv:2305.02463}, 2023.

\bibitem{liang2022self}
Hanwen Liang, Niamul Quader, Zhixiang Chi, Lizhe Chen, Peng Dai, Juwei Lu, and Yang Wang.
\newblock Self-supervised spatiotemporal representation learning by exploiting video continuity.
\newblock In {\em Proceedings of the AAAI Conference on Artificial Intelligence}, volume~36, pages 1564--1573, 2022.

\bibitem{lin2023magic3d}
Chen-Hsuan Lin, Jun Gao, Luming Tang, Towaki Takikawa, Xiaohui Zeng, Xun Huang, Karsten Kreis, Sanja Fidler, Ming-Yu Liu, and Tsung-Yi Lin.
\newblock Magic3d: High-resolution text-to-3d content creation.
\newblock In {\em Proceedings of the IEEE/CVF Conference on Computer Vision and Pattern Recognition}, pages 300--309, 2023.

\bibitem{liu2023zero}
Ruoshi Liu, Rundi Wu, Basile Van~Hoorick, Pavel Tokmakov, Sergey Zakharov, and Carl Vondrick.
\newblock Zero-1-to-3: Zero-shot one image to 3d object.
\newblock In {\em Proceedings of the IEEE/CVF International Conference on Computer Vision}, pages 9298--9309, 2023.

\bibitem{liu2023syncdreamer}
Yuan Liu, Cheng Lin, Zijiao Zeng, Xiaoxiao Long, Lingjie Liu, Taku Komura, and Wenping Wang.
\newblock Syncdreamer: Generating multiview-consistent images from a single-view image.
\newblock {\em arXiv preprint arXiv:2309.03453}, 2023.

\bibitem{long2023wonder3d}
Xiaoxiao Long, Yuan-Chen Guo, Cheng Lin, Yuan Liu, Zhiyang Dou, Lingjie Liu, Yuexin Ma, Song-Hai Zhang, Marc Habermann, Christian Theobalt, et~al.
\newblock Wonder3d: Single image to 3d using cross-domain diffusion.
\newblock {\em arXiv preprint arXiv:2310.15008}, 2023.

\bibitem{lu2023direct2}
Yuanxun Lu, Jingyang Zhang, Shiwei Li, Tian Fang, David McKinnon, Yanghai Tsin, Long Quan, Xun Cao, and Yao Yao.
\newblock Direct2. 5: Diverse text-to-3d generation via multi-view 2.5 d diffusion.
\newblock {\em arXiv preprint arXiv:2311.15980}, 2023.

\bibitem{luo2024scalable}
Tiange Luo, Chris Rockwell, Honglak Lee, and Justin Johnson.
\newblock Scalable 3d captioning with pretrained models.
\newblock {\em Advances in Neural Information Processing Systems}, 36, 2024.

\bibitem{nichol2021glide}
Alex Nichol, Prafulla Dhariwal, Aditya Ramesh, Pranav Shyam, Pamela Mishkin, Bob McGrew, Ilya Sutskever, and Mark Chen.
\newblock Glide: Towards photorealistic image generation and editing with text-guided diffusion models.
\newblock {\em arXiv preprint arXiv:2112.10741}, 2021.

\bibitem{nichol2022point}
Alex Nichol, Heewoo Jun, Prafulla Dhariwal, Pamela Mishkin, and Mark Chen.
\newblock Point-e: A system for generating 3d point clouds from complex prompts.
\newblock {\em arXiv preprint arXiv:2212.08751}, 2022.

\bibitem{niemeyer2022regnerf}
Michael Niemeyer, Jonathan~T Barron, Ben Mildenhall, Mehdi~SM Sajjadi, Andreas Geiger, and Noha Radwan.
\newblock Regnerf: Regularizing neural radiance fields for view synthesis from sparse inputs.
\newblock In {\em Proceedings of the IEEE/CVF Conference on Computer Vision and Pattern Recognition}, pages 5480--5490, 2022.

\bibitem{pan2023enhancing}
Zijie Pan, Jiachen Lu, Xiatian Zhu, and Li~Zhang.
\newblock Enhancing high-resolution 3d generation through pixel-wise gradient clipping.
\newblock {\em arXiv preprint arXiv:2310.12474}, 2023.

\bibitem{pan2024fast}
Zijie Pan, Zeyu Yang, Xiatian Zhu, and Li~Zhang.
\newblock Fast dynamic 3d object generation from a single-view video.
\newblock {\em arXiv preprint arXiv:2401.08742}, 2024.

\bibitem{poole2022dreamfusion}
Ben Poole, Ajay Jain, Jonathan~T Barron, and Ben Mildenhall.
\newblock Dreamfusion: Text-to-3d using 2d diffusion.
\newblock {\em arXiv preprint arXiv:2209.14988}, 2022.

\bibitem{radford2021learning}
Alec Radford, Jong~Wook Kim, Chris Hallacy, Aditya Ramesh, Gabriel Goh, Sandhini Agarwal, Girish Sastry, Amanda Askell, Pamela Mishkin, Jack Clark, et~al.
\newblock Learning transferable visual models from natural language supervision.
\newblock In {\em International conference on machine learning}, pages 8748--8763. PMLR, 2021.

\bibitem{ren2023dreamgaussian4d}
Jiawei Ren, Liang Pan, Jiaxiang Tang, Chi Zhang, Ang Cao, Gang Zeng, and Ziwei Liu.
\newblock Dreamgaussian4d: Generative 4d gaussian splatting.
\newblock {\em arXiv preprint arXiv:2312.17142}, 2023.

\bibitem{rombach2022high}
Robin Rombach, Andreas Blattmann, Dominik Lorenz, Patrick Esser, and Bj{\"o}rn Ommer.
\newblock High-resolution image synthesis with latent diffusion models.
\newblock In {\em Proceedings of the IEEE/CVF conference on computer vision and pattern recognition}, pages 10684--10695, 2022.

\bibitem{ruiz2023dreambooth}
Nataniel Ruiz, Yuanzhen Li, Varun Jampani, Yael Pritch, Michael Rubinstein, and Kfir Aberman.
\newblock Dreambooth: Fine tuning text-to-image diffusion models for subject-driven generation.
\newblock In {\em Proceedings of the IEEE/CVF Conference on Computer Vision and Pattern Recognition}, pages 22500--22510, 2023.

\bibitem{saito2017temporal}
Masaki Saito, Eiichi Matsumoto, and Shunta Saito.
\newblock Temporal generative adversarial nets with singular value clipping.
\newblock In {\em Proceedings of the IEEE international conference on computer vision}, pages 2830--2839, 2017.

\bibitem{shi2023zero123++}
Ruoxi Shi, Hansheng Chen, Zhuoyang Zhang, Minghua Liu, Chao Xu, Xinyue Wei, Linghao Chen, Chong Zeng, and Hao Su.
\newblock Zero123++: a single image to consistent multi-view diffusion base model.
\newblock {\em arXiv preprint arXiv:2310.15110}, 2023.

\bibitem{shi2023mvdream}
Yichun Shi, Peng Wang, Jianglong Ye, Mai Long, Kejie Li, and Xiao Yang.
\newblock Mvdream: Multi-view diffusion for 3d generation.
\newblock {\em arXiv preprint arXiv:2308.16512}, 2023.

\bibitem{singer2022make}
Uriel Singer, Adam Polyak, Thomas Hayes, Xi~Yin, Jie An, Songyang Zhang, Qiyuan Hu, Harry Yang, Oron Ashual, Oran Gafni, et~al.
\newblock Make-a-video: Text-to-video generation without text-video data.
\newblock {\em arXiv preprint arXiv:2209.14792}, 2022.

\bibitem{singer2023text}
Uriel Singer, Shelly Sheynin, Adam Polyak, Oron Ashual, Iurii Makarov, Filippos Kokkinos, Naman Goyal, Andrea Vedaldi, Devi Parikh, Justin Johnson, et~al.
\newblock Text-to-4d dynamic scene generation.
\newblock {\em arXiv preprint arXiv:2301.11280}, 2023.

\bibitem{song2020denoising}
Jiaming Song, Chenlin Meng, and Stefano Ermon.
\newblock Denoising diffusion implicit models.
\newblock {\em arXiv preprint arXiv:2010.02502}, 2020.

\bibitem{tang2024lgm}
Jiaxiang Tang, Zhaoxi Chen, Xiaokang Chen, Tengfei Wang, Gang Zeng, and Ziwei Liu.
\newblock Lgm: Large multi-view gaussian model for high-resolution 3d content creation.
\newblock {\em arXiv preprint arXiv:2402.05054}, 2024.

\bibitem{tang2023dreamgaussian}
Jiaxiang Tang, Jiawei Ren, Hang Zhou, Ziwei Liu, and Gang Zeng.
\newblock Dreamgaussian: Generative gaussian splatting for efficient 3d content creation.
\newblock {\em arXiv preprint arXiv:2309.16653}, 2023.

\bibitem{tang2024mvdiffusion++}
Shitao Tang, Jiacheng Chen, Dilin Wang, Chengzhou Tang, Fuyang Zhang, Yuchen Fan, Vikas Chandra, Yasutaka Furukawa, and Rakesh Ranjan.
\newblock Mvdiffusion++: A dense high-resolution multi-view diffusion model for single or sparse-view 3d object reconstruction.
\newblock {\em arXiv preprint arXiv:2402.12712}, 2024.

\bibitem{unterthiner2019fvd}
Thomas Unterthiner, Sjoerd van Steenkiste, Karol Kurach, Rapha{\"e}l Marinier, Marcin Michalski, and Sylvain Gelly.
\newblock Fvd: A new metric for video generation.
\newblock 2019.

\bibitem{voleti2024sv3d}
Vikram Voleti, Chun-Han Yao, Mark Boss, Adam Letts, David Pankratz, Dmitry Tochilkin, Christian Laforte, Robin Rombach, and Varun Jampani.
\newblock Sv3d: Novel multi-view synthesis and 3d generation from a single image using latent video diffusion.
\newblock {\em arXiv preprint arXiv:2403.12008}, 2024.

\bibitem{vondrick2016generating}
Carl Vondrick, Hamed Pirsiavash, and Antonio Torralba.
\newblock Generating videos with scene dynamics.
\newblock {\em Advances in neural information processing systems}, 29, 2016.

\bibitem{wang2023modelscope}
Jiuniu Wang, Hangjie Yuan, Dayou Chen, Yingya Zhang, Xiang Wang, and Shiwei Zhang.
\newblock Modelscope text-to-video technical report.
\newblock {\em arXiv preprint arXiv:2308.06571}, 2023.

\bibitem{wang2023imagedream}
Peng Wang and Yichun Shi.
\newblock Imagedream: Image-prompt multi-view diffusion for 3d generation.
\newblock {\em arXiv preprint arXiv:2312.02201}, 2023.

\bibitem{wang2024prolificdreamer}
Zhengyi Wang, Cheng Lu, Yikai Wang, Fan Bao, Chongxuan Li, Hang Su, and Jun Zhu.
\newblock Prolificdreamer: High-fidelity and diverse text-to-3d generation with variational score distillation.
\newblock {\em Advances in Neural Information Processing Systems}, 36, 2024.

\bibitem{wang2004image}
Zhou Wang, Alan~C Bovik, Hamid~R Sheikh, and Eero~P Simoncelli.
\newblock Image quality assessment: from error visibility to structural similarity.
\newblock {\em IEEE transactions on image processing}, 13(4):600--612, 2004.

\bibitem{wu20234d}
Guanjun Wu, Taoran Yi, Jiemin Fang, Lingxi Xie, Xiaopeng Zhang, Wei Wei, Wenyu Liu, Qi~Tian, and Xinggang Wang.
\newblock 4d gaussian splatting for real-time dynamic scene rendering.
\newblock {\em arXiv preprint arXiv:2310.08528}, 2023.

\bibitem{xu2024comp4d}
Dejia Xu, Hanwen Liang, Neel~P Bhatt, Hezhen Hu, Hanxue Liang, Konstantinos~N Plataniotis, and Zhangyang Wang.
\newblock Comp4d: Llm-guided compositional 4d scene generation.
\newblock {\em arXiv preprint arXiv:2403.16993}, 2024.

\bibitem{yang2024diffusion}
Zeyu Yang, Zijie Pan, Chun Gu, and Li~Zhang.
\newblock Diffusion$^2$: Dynamic 3d content generation via score composition of orthogonal diffusion models.
\newblock {\em arXiv preprint arXiv:2404.02148}, 2024.

\bibitem{yi2023gaussiandreamer}
Taoran Yi, Jiemin Fang, Guanjun Wu, Lingxi Xie, Xiaopeng Zhang, Wenyu Liu, Qi~Tian, and Xinggang Wang.
\newblock Gaussiandreamer: Fast generation from text to 3d gaussian splatting with point cloud priors.
\newblock {\em arXiv preprint arXiv:2310.08529}, 2023.

\bibitem{yin20234dgen}
Yuyang Yin, Dejia Xu, Zhangyang Wang, Yao Zhao, and Yunchao Wei.
\newblock 4dgen: Grounded 4d content generation with spatial-temporal consistency.
\newblock {\em arXiv preprint arXiv:2312.17225}, 2023.

\bibitem{yu2023video}
Sihyun Yu, Kihyuk Sohn, Subin Kim, and Jinwoo Shin.
\newblock Video probabilistic diffusion models in projected latent space.
\newblock In {\em Proceedings of the IEEE/CVF Conference on Computer Vision and Pattern Recognition}, pages 18456--18466, 2023.

\bibitem{zeng2024stag4d}
Yifei Zeng, Yanqin Jiang, Siyu Zhu, Yuanxun Lu, Youtian Lin, Hao Zhu, Weiming Hu, Xun Cao, and Yao Yao.
\newblock Stag4d: Spatial-temporal anchored generative 4d gaussians.
\newblock {\em arXiv preprint arXiv:2403.14939}, 2024.

\bibitem{zhang2018unreasonable}
Richard Zhang, Phillip Isola, Alexei~A Efros, Eli Shechtman, and Oliver Wang.
\newblock The unreasonable effectiveness of deep features as a perceptual metric.
\newblock In {\em Proceedings of the IEEE conference on computer vision and pattern recognition}, pages 586--595, 2018.

\bibitem{zhang2023i2vgen}
Shiwei Zhang, Jiayu Wang, Yingya Zhang, Kang Zhao, Hangjie Yuan, Zhiwu Qin, Xiang Wang, Deli Zhao, and Jingren Zhou.
\newblock I2vgen-xl: High-quality image-to-video synthesis via cascaded diffusion models.
\newblock {\em arXiv preprint arXiv:2311.04145}, 2023.

\bibitem{zhao2023animate124}
Yuyang Zhao, Zhiwen Yan, Enze Xie, Lanqing Hong, Zhenguo Li, and Gim~Hee Lee.
\newblock Animate124: Animating one image to 4d dynamic scene.
\newblock {\em arXiv preprint arXiv:2311.14603}, 2023.

\bibitem{zuo2024videomv}
Qi~Zuo, Xiaodong Gu, Lingteng Qiu, Yuan Dong, Zhengyi Zhao, Weihao Yuan, Rui Peng, Siyu Zhu, Zilong Dong, Liefeng Bo, et~al.
\newblock Videomv: Consistent multi-view generation based on large video generative model.
\newblock {\em arXiv preprint arXiv:2403.12010}, 2024.

\end{thebibliography}

\newpage
\appendix

\section{Appendix}

\vspace{-0.2cm}
\subsection{Related Work}
\label{app:related_work}

\vspace{-0.2cm}
\subsubsection{3D Generation}
3D generation aims to create static 3D content from text or images. DreamFusion~\cite{poole2022dreamfusion}, first introduces the score distillation
sampling (SDS) loss to optimize NeRF with diffusion models. However, The original form of SDS encounters challenges such as multi-face Janus issues and slow optimization. Subsequent works \cite{liu2023zero,liu2023syncdreamer,wang2023imagedream,wang2024prolificdreamer,pan2023enhancing,shi2023zero123++,shi2023mvdream,yi2023gaussiandreamer} have attempted to refine the mechanism to address these issues. Another research direction \cite{liu2023zero,long2023wonder3d,liu2023syncdreamer,tang2024mvdiffusion++} focus on generating dense multi-view images with consistent 3D properties directly. They achieve this by training or fine-tuning 2D diffusion models on 3D datasets to better suit 3D generation tasks. Zero123 \cite{liu2023zero}  generates a 2D image from an unseen view based on a single image.  SyncDreamer \cite{liu2023syncdreamer} can generate multiview-consistent images from a single input image. Some later works \cite{shi2023zero123++,shi2023mvdream} explicitly generate fixed multi-view images in one diffusion pass. The generated
images can be used for reconstructing 3D content like implicit radiance field, point cloud, or textured meshes. Some other works like Point-E \cite{nichol2022point} and Shap-E \cite{jun2023shap} train models to directly generate 3D point clouds or meshes. Except for equipping the model multi-view aware ability, some works \cite{lin2023magic3d,lu2023direct2,tang2023dreamgaussian} focus on 3D representations study to improve efficiency and generation performance.

\subsubsection{Video Generation}
Video generation has gained significant attention in recent years. Previous works \cite{vondrick2016generating,saito2017temporal} adopted GANs to directly learn the
joint distribution of video frames. Diffusion has been employed for video prediction
tasks in recent works, which have achieved great
levels of realism, diversity, and controllability. Among them, video LDM \cite{blattmann2023align} applies the latent diffusion framework for video generation. SVD \cite{blattmann2023stable} follows the same architecture and
inserts temporal convolution and attention layers after every
spatial convolution and attention layer, achieving effective performance improvement. W.A.L.T \cite{gupta2023photorealistic}  introduces window attention architecture for
joint spatial and spatio-temporal generative modeling. Some recent works \cite{voleti2024sv3d,zuo2024videomv} have tried to leverage temporal consistency~\cite{liang2022self} from video generation but are limited to static 3D generation tasks. Inspired by their success, our work focuses on utilizing spatial consistency from video diffusion models for 4D generation tasks.

\subsubsection{4D Generation}
 Compared to 3D generation, 4D generation requires not only predicting consistent geometry but also generating temporal-consistent dynamics. There are a few recent works dedicated to this task. One line of research works predict 4D representations conditioned on a single image and text description. MAV3D \cite{singer2023text} deals with a text-to-4D problem by utilizing score distillation sampling derived from video diffusion
models to optimize a dynamic NeRF based on textual prompts. Animate124 \cite{zhao2023animate124} leverages a dynamic NeRF-based representation to tackle this problem. 4DFY \cite{bahmani20234d} addresses text-to-4D synthesis by combining supervision signals from image, video and 3D-aware diffusion models. DreamGaussian4D \cite{ren2023dreamgaussian4d} adopts
the deformable 3D Gaussian as 4D representations. 4DGen~\cite{yin20234dgen} proposes a novel pipeline where they utilize spatial-temporal pseudo labels into anchor frames with a multi-view diffusion model. Another line of work predicts dynamic objects from single-view videos. Consistent4D \cite{jiang2023consistent4d} leverages the object-level 3D-aware image diffusion model as the primary supervision signal for training Dynamic Neural Radiance Fields, and proposing cascade DyNeRF to facilitate stable training. More recent work STAG4D \cite{zeng2024stag4d} utilizes a multi-view diffusion model to initialize multi-view images anchoring on the input video frames and introduce a fusion strategy to ensure temporal consistency. Efficient4D \cite{pan2024fast} generates multi-frame multi-view captures of dynamic 3D objects through SyncDreamer-T and reconstructs 4D representations with them. Different from previous methods, our framework is the first to embed spatial-temporal consistency into a singular video diffusion model to achieve efficient and scalable generation of 4D content.

\begin{figure}
    \begin{center}
        \includegraphics[width=1.0\linewidth]
        {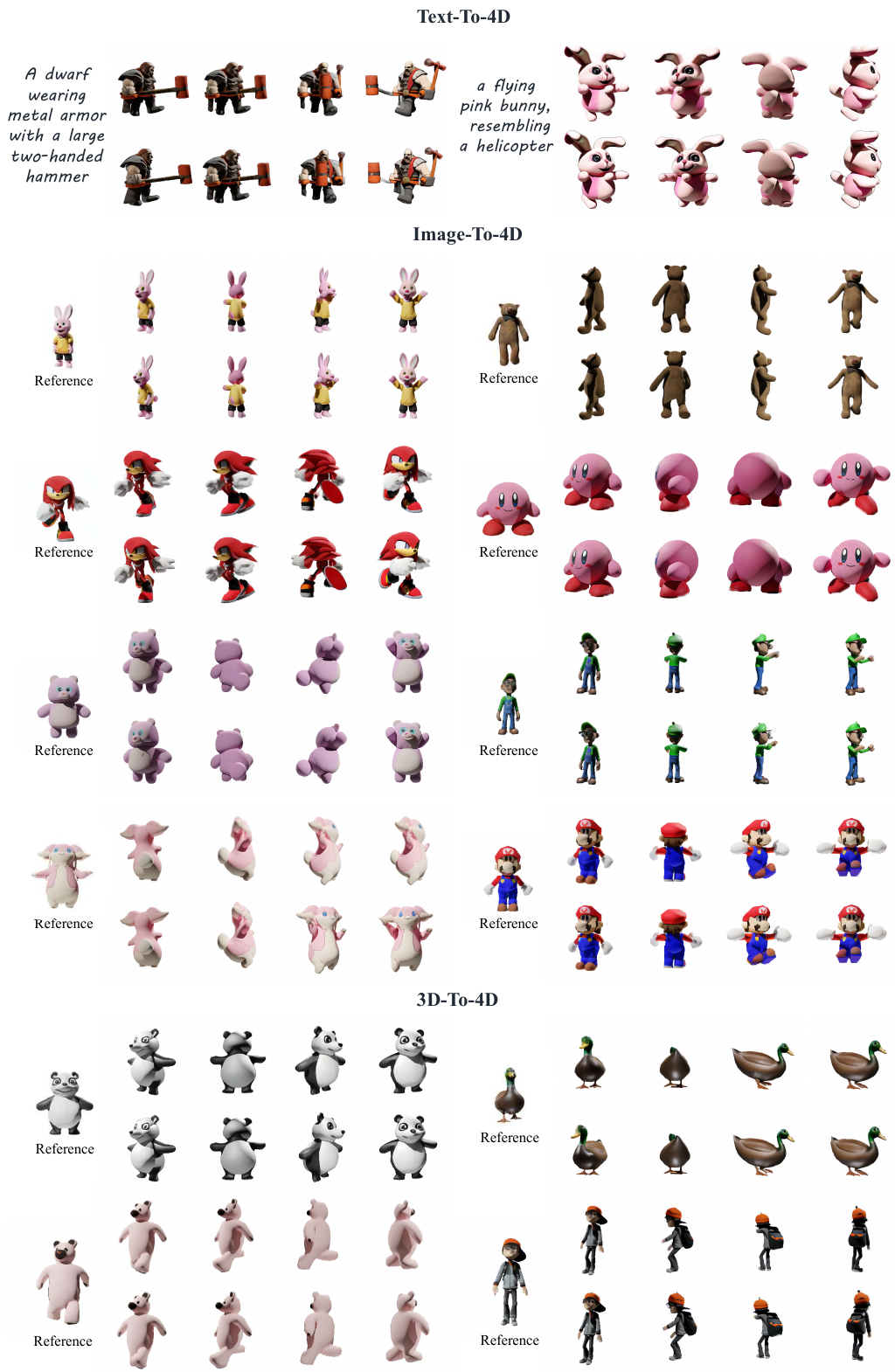}
        % \vspace{-0.2cm}
        \caption{Visualization of more experimental results. For each prompt, the first row indicates results from 4D construction, and the second row indicates results from 4D-aware diffusion model.}
        \label{fig:app}
    \end{center}
    % \vspace{-0.3cm}
\end{figure}

\vspace{-0.2cm}
\subsection{More Experimental Results}
We visualize more experimental results in Fig.~\ref{fig:app}. For each task, with each prompt, we show 4 views from consecutive timestamps to demonstrate the spatial-temporal motions and the 3D geometry consistency. As before, we show in the first row the results from 4D construction, and in the second row the results from 4D-aware diffusion model.

\subsection{More Experimental Details}
\label{app:exp_detail}
Here we provide more experimental details about our methods as well as the baselines.
We train our 4D-aware video diffusion model for 4k iterations with a constant learning rate of $3\times 10^{-5}$. We use the AdamW optimizer and employ FP16 for efficient gradient descent without weight decay. The motion magnitude reconstruction loss weight $\omega$ is set to be $5\times 10^{-4}$. We optimized the model on 4 NVIDIA A100 GPUs. We use a total batch size of 64, with a batch size of 16 on each GPU. The training procedure requires roughly 80GB of VRAM and it takes around 16 hours. 
In the 4D construction stage, the input comprises 24-frame static multi-view images produced by VideoMV and 24-frame dynamic multi-view images. We optimized the 4DGS representation for 5,000 iterations in the coarse stage and 2,000 iterations in the fine stage. The hexplane resolution is set at $64 \times64 \times64 \times24$. The loss weights assigned to $L_1$, $L_{lpips}$ and $L_{depeth}$ are 1,10 and 1 respectively. The other hyperparameter settings (including the learning rates and optimizer) are the same as 4DGS\cite{wu20234d}.

For the baselines, we followed the instructions from their public websites to do experiments. 4DFY~\cite{bahmani20234d} is designed for text-to-4D generation. We fed in the text prompts directly to get the generation results. 
Animate124~\cite{zhao2023animate124} requires a pair of text and image prompts for generation. Therefore, we use the first front-view image in our generated video as the image prompt for generation. In the image-to-4D generation, both 4DGen~\cite{yin20234dgen} and STAG4D~\cite{zeng2024stag4d} propose to use pretrained video diffusion models to generate monocular videos as the prompt conditions. To boost their performance, we directly render ground-truth images at the front view from the source dynamic 3D datasets and obtain the monocular dynamic videos as prompts.

\subsection{Limitation and Future work}
\label{app:limitation}
While our proposed Diffusion4D framework demonstrates significant advancements in terms of efficiency and consistency for 4D content generation, we acknowledge the following limitations to provide a foundation for future work and improvements.
We are leveraging VideoMV~\cite{zuo2024videomv} as our base 3D-aware video diffusion model, with a video size of $24\times 256 \times256$. While this resolution provides a good balance between quality and computational feasibility, higher resolution and longer temporal sequences are needed for more detailed and realistic 4D content. Increasing the image resolution by training on larger-size rendered images or using a super-resolution model could enhance the visualization effect on 4D construction. Similarly, increasing the temporal dimension to obtain denser views of the dynamic 3D objects will improve the fidelity of 3D geometry details.
Our future research will focus on expanding the diversity and quality of the dynamic 3D dataset. We will also explore extending the framework to handle other conditions, such as single-view video, which can further broaden the applications of Diffusion4D. 
We will put more effort in generating longer and high-resolution 4D orbital videos to facilitate better 4D construction.

\end{document}